\documentclass[10pt, twocolumn, journal]{IEEEtran}
\usepackage{amsmath}
\usepackage[]{graphicx}
\usepackage{subfig}
\usepackage[]{algorithm2e}
\usepackage[]{algorithmic}
\usepackage{color}
\usepackage{soul}
\usepackage{booktabs} 
\usepackage{hyperref}
\usepackage{float}

\def \figscale {0.95}

\newcommand{\uCT}{$\mu$CT}
\def \Tforward {\mathbf{T}_\mathrm{forward}}
\def \Tbackward {\mathbf{T}_\mathrm{backward}}

\newcommand{\ifmiaAuthorAff}[2]{\vspace{1mm} ${\,}^{#1}$ \normalsize{\it{#2}}}

\newcommand{\ifmiaKeywords}[1]{\vspace{1mm} \noindent \small{{\bf Keywords: } #1}}

\begin{document}

\title{\Large{Multi-scale Image Fusion Between Pre-operative Clinical CT \\ and X-ray Microtomography of Lung Pathology}}

\author{Holger R. Roth\ifmiaAuthorAff{a}, Kai Nagara\ifmiaAuthorAff{b}, Hirohisa Oda\ifmiaAuthorAff{b}, \\ 
Masahiro Oda\ifmiaAuthorAff{b}, Tomoshi Sugiyama\ifmiaAuthorAff{c}, Shota Nakamura\ifmiaAuthorAff{c}, Kensaku Mori\ifmiaAuthorAff{a,b}\\

\ifmiaAuthorAff{a}{Information \& Communications, Nagoya University, Japan}\\
\ifmiaAuthorAff{b}{Graduate School of Information Science, Nagoya University, Japan}\\
\ifmiaAuthorAff{c}{Nagoya University Graduate School of Medicine, Japan}
\vspace{-7mm}}


\maketitle

\begin{abstract}
Computational anatomy allows the quantitative analysis of organs in medical images. However, most analysis is constrained to the millimeter scale because of the limited resolution of clinical computed tomography (CT). X-ray microtomography (\uCT{}) on the other hand allows imaging of ex-vivo tissues at a resolution of tens of microns. In this work, we use clinical CT to image lung cancer patients before partial pneumonectomy (resection of pathological lung tissue). The resected specimen is prepared for \uCT{} imaging at a voxel resolution of 50 $\mu$m (0.05 mm). This high-resolution image of the lung cancer tissue allows further insides into understanding of tumor growth and categorization. For making full use of this additional information, image fusion (registration) needs to be performed in order to re-align the \uCT{} image with clinical CT. We developed a multi-scale non-rigid registration approach. After manual initialization using a few landmark points and rigid alignment, several levels of non-rigid registration between down-sampled (in the case of  \uCT{}) and up-sampled (in the case of clinical CT) representations of the image are performed. Any non-lung tissue is ignored during the computation of the similarity measure used to guide the registration during optimization. We are able to recover the volume differences introduced by the resection and preparation of the lung specimen. The average ($\pm$ std. dev.) minimum surface distance between \uCT{} and clinical CT at the resected lung surface is reduced from 3.3 $\pm$ 2.9 (range: [0.1, 15.9]) to 2.3 mm $\pm$ 2.8 (range: [0.0, 15.3]) mm. This is a significant improvement with p $<$ 0.001 (Wilcoxon Signed Rank Test). The alignment of clinical CT with \uCT{} will allow further registration with even finer resolutions of \uCT{} (up to 10 $\mu$m resolution) and ultimately with histopathological microscopy images for further macro to micro image fusion that can aid medical image analysis.
\end{abstract}

\ifmiaKeywords{image fusion, non-rigid registration, computed tomography (CT), X-ray microtomography (\uCT{})}

\section{INTRODUCTION}
\label{sec:intro}
\noindent Clinical computed tomography (CT) is used for the diagnostic imaging of the living human (in-vivo imaging). As a result, most computational analysis is constrained to the millimeter scale because of the limited resolution of clinical CT. At this millimeter scale, pulmonary blood vessels and lung lobes can be observed. However, finer detailed anatomy is not observable. X-ray microtomograph (\uCT) on the other hand allows imaging of ex-vivo tissues at a resolution of tens of microns. At this $\mu$m-scale, the alveoli and bronchiole regions can be clearly observed \cite{mori2016macro}.

The prospective study of pre-operative imaging together with the high-resolution image analysis of resected tissue after surgery may provide us with unique opportunities to verify and potentially improve imaging protocols for cancer diagnostics \cite{goubran2013image,eriksson2007correlation, howe2010histologically}. 

In this work, we use clinical CT to image lung cancer patients before partial pneumonectomy (resection of pathological lung tissue). The resected specimen is then prepared for \uCT{} imaging allowing the imaging of $\mu$m-scale anatomy. This study investigates the use of non-rigid intensity-based registration in order to establish a scale-seamless registration between clinical CT and \uCT{}, with the ultimate aim of allowing a seamless navigation between anatomical scales inside the human body \cite{mori2016macro}. Related work is the registration of in-vivo and ex-vivo MRI of surgically resected specimens by \cite{goubran2015registration,goubran2013image}.
\section{METHOD}
\label{sec:method}
\noindent The resected specimen is prepared for \uCT{} imaging at a voxel resolution of 50 $\mu$m (0.05 mm). This high-resolution image of the lung cancer tissue allows further insides into understanding of tumor growth and categorization. For making full use of this additional information, image fusion (registration) needs to be performed in order to re-align the \uCT{} image with clinical CT. We developed a multi-scale non-rigid registration approach. 
\subsection{Establishing scale-seamless registration between \uCT{} and clinical CT}
\label{sec:establish_correspondence}
We use a non-rigid registration method to align the \uCT{} specimen with the pre-operative clinical CT. This method is driven by the intensity similarity between the images. Any non-lung tissue is ignored during computation of the similarity measure $\mathcal{S}$ used to guide the registration during optimization. This is achieved via simple thresholding and morphological operations to extract the lung region in clinical CT and \uCT{} as pre-processing step. 

A coarse-to-fine approach is proposed in order to capture first the largest deformations and then the smaller differences between both images. This is achieved with a four-level multi-resolution pyramid as illustrated in Fig. \ref{fig:uct_to_cct}. In order to compensate for the large differences in resolution between the \uCT{} (in the case of  \uCT{}) and clinical CT images, we use a \textit{down-sample} and \textit{up-sample} (in the case of clinical CT) pyramid and perform the registration optimizations at each level in a coarse-to-fine fashion.
\begin{figure}[htb]
  \centering
   \includegraphics[width=\figscale\columnwidth, clip]{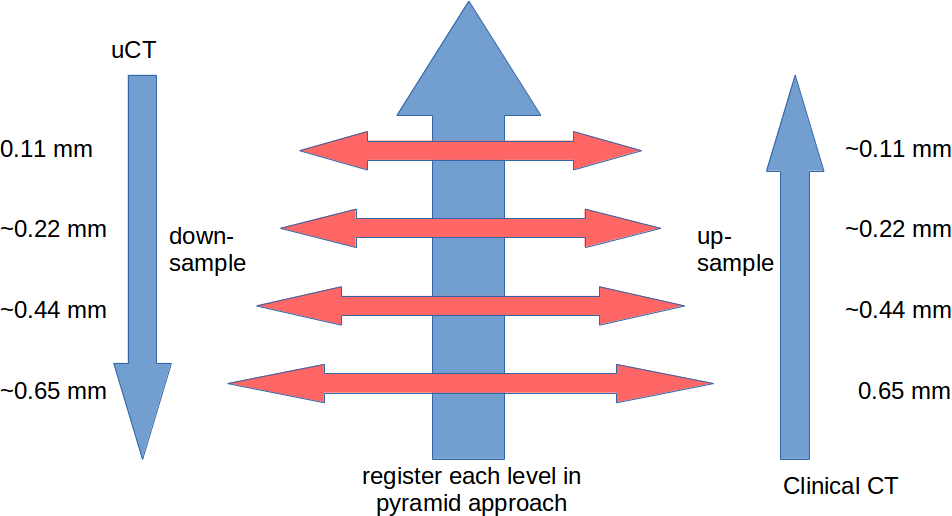}
   \caption{Multi-scale image fusion between pre-operative clinical CT and x-ray microtomography of lung pathology in a coarse-to-fine approach.}
   \label{fig:uct_to_cct}
\end{figure}
A good initialization for the registration algorithm is important. We use a few (3$\sim$5) manual point in order to provide affine alignment of the specimen with the pre-operative CT image. After affine initialization, we establish non-rigid alignment using the B-spline registration method, also known as 3D free form deformation based registration of \cite{Rueckert1999nonrigid} with the implementation provided by \cite{modat2010fast}.

A 3D cubic B-Splines deformation model uses a lattice of control points $\{\vec{\phi}\}$. The spacing between each control point is uniform and denoted as $\delta_x$, $\delta_y$, and $\delta_z$ along the $x$-, $y$-, and $z$-axis respectively. For each voxel $\vec{x}$ in the domain $\Omega$ of the target image. the deformation $\mathbf{T}(\vec{x})$ can be computed as:
\begin{equation}
  \label{equ:T_3D}
	\mathbf{T}(\vec{x}) = \sum_{i,j,k}{ \beta^3(\frac{x}{\delta_x} - i) \times \beta^3(\frac{y}{\delta_y} - j) \times \beta^3(\frac{y}{\delta_z} - k) \times \vec{\phi}_{ijk} },
\end{equation} 
where $\beta^3$ represents the cubic B-Spline function.

The images are aligned by finding the transformation which maximizes the following objective function:
\begin{multline} 
	\mathcal{O}\left(I_\mathrm{p},I_\mathrm{s}\left(\mathbf{T}\right);\{\vec{\phi}\}\right) = 	\left(1 - \alpha - \beta - \gamma\right)\times\mathcal{S} \\ 
	- {\alpha}\times\mathcal{C}_\mathrm{smooth}(\mathbf{T}) - {\beta}\times\mathcal{C}_\mathrm{volpres}(\mathbf{T}) -{\gamma}\times\mathcal{C}_\mathrm{inconsistency}(\mathbf{T})
\end{multline}
which combines a similarity measure, $\mathcal{S}$, and three penalty constraint terms, $\mathcal{C}_\mathrm{smooth}$, $\mathcal{C}_\mathrm{volpres}$, and $\mathcal{C}_\mathrm{inconsistency}$. Each term is weighted against each other by user-defined weights $\alpha$, $\beta$, and $\gamma$. 

The similarity measure used between the reference ($R$) and floating image ($F$) is normalized mutual information (NMI): 
\begin{equation} 
	\mathcal{S} \equiv \mathbf{NMI} = \frac{   H(R)  + H(F(\mathbf{T}))  }{   H(R, F(\mathbf{T})) }
\end{equation} 
where $H(R)$ and $H(F(\mathbf{T}))$ the two marginal entropies, and $H(R, F(\mathbf{T})$) is the joint entropy. Its computation requires a joint histogram which is filled by using a Parzen Window (PW) approach \cite{modat2010fast,mattes2003pet}.

The three constraint terms are used to encourage realistic deformations. The bending energy describes the smoothness of the deformation and is defined as:
\begin{multline} 
	\mathcal{C}_\mathrm{smooth} =
	 \frac{1}{N} \sum_{\vec{x}\forall\Omega}(
	 \left|\frac{\partial^2\mathbf{T}\left(\vec{x}\right)}{{\partial}x^2}\right|^2 +
        \left|\frac{\partial^2\mathbf{T}\left(\vec{x}\right)}{{\partial}y^2}\right|^2 + 
        \left|\frac{\partial^2\mathbf{T}\left(\vec{x}\right)}{{\partial}z^2}\right|^2 \\
+ 2\times\left[\left|\frac{\partial^2\mathbf{T}\left(\vec{x}\right)}{{\partial}xy}\right|^2 +
                \left|\frac{\partial^2\mathbf{T}\left(\vec{x}\right)}{{\partial}yz}\right|^2 +
                \left|\frac{\partial^2\mathbf{T}\left(\vec{x}\right)}{{\partial}xz}\right|^2 
     \right])
.
\end{multline} 
\noindent The volume-preserving penalty term discourages large expansions/contractions, and is defined as:
\begin{equation} 
	\mathcal{C}_\mathrm{volpres} = \frac{1}{N}\sum_{\vec{x}\forall\Omega}
		\left[
                          \log
                                \left(
                                          \det{
                                                   \left(
                                                            \mathrm{Jac}\left(\mathbf{T}\left(\vec{x}\right)\right)
                                                   \right)
                                                   }
                               \right)
             \right]^2
\end{equation}
In addition we prevent the occurrence of folding in the transformation using a folding correction scheme \cite{modat2010lung}. For each transformed voxel that would cause a negative Jacobian determinant, its influence on its neighborhood control points is computed. The control point positions is then changed until the determinant value is positive. 

In an ideal case, the transformations from $F$ to $R$ (forward) and $R$ to $F$ (backward) are the inverse of each other, e.g.  $\Tforward=\Tbackward^{-1}$ and $\Tbackward=\Tforward^{-1}$ \cite{christensen2006introduction}. Hence, we include a penalty term that encourages inverse consistency of both transformations. We follow the approach of \cite{feng2009new} using compositions of $\Tforward$ and $\Tbackward$ and add 
\begin{multline}
	\label{inv_const_penalty}
	\mathcal{C}_\mathrm{inconsistency} = 
		\sum_{\vec{x}\forall \Omega}\left\|\Tforward\left(\Tbackward\left(\vec{x}\right)\right)\right\|^2 \\
	+ \sum_{\vec{x}\forall \Omega}\left\|\Tbackward\left(\Tforward\left(\vec{x}\right)\right)\right\|^2
\end{multline}
The following parameters were found empirically by visual examination of the registration results. We use a four-level multi-resolution pyramid with a maximum of 500 iterations per level. Both the image and B-spline control point grid resolutions are doubled with increasing resolution levels. The final control point spacing between voxels is 5. The objective function weights are set to $\alpha = 10^{-4}$, $\beta = 10^{-12}$, and $\gamma = 0.1$. These parameters were found to recover the majority of the deformation between the two images, while preventing unrealistic deformations from occurring. We used the open-source software\footnote{\url{http://sourceforge.net/projects/niftyreg}} by \cite{modat2010fast,modat2012inverse} for this study.
\section{RESULTS}
\label{sec:results}
\noindent We collected pre-operative clinical CTs from two patients before partial pneumonectomy. The clinically acquired CT images have dimensions of [512, 512, 435$\sim$554], and voxel spacings of [0.625, 0.625, 0.6] mm. The \uCT{} images used for registration had dimensions of [278$\sim$512, 278$\sim$512, 346$\sim$538] with  isotropic voxel spacings of $0.111\sim0.127$ mm. Note that we downsampled the original \uCT{} images (circa 50$\mu$m resolution) by a factor of two for the registration experiments. The tube voltage for \uCT{} was 90kVp, and tube current was 110 $\mu$A.

3$\sim$5 corresponding points where chosen by an expert clinician (SN) in order to provide an initial affine alignment of the specimen with the pre-operative CT image. We then crop the clinical CT image to the extent of the aligned \uCT{} image for subsequent non-rigid registration.

Figure \ref{fig:axcosa} shows the alignment of \uCT{} lung specimen with pre-operative clinical CT before and after non-rigid registration for one case. A qualitatively better alignment of \uCT{} with the target clinical CT regions can be observed. 
\begin{figure}[htb]
  \centering
   \includegraphics[width=\figscale\columnwidth, clip]{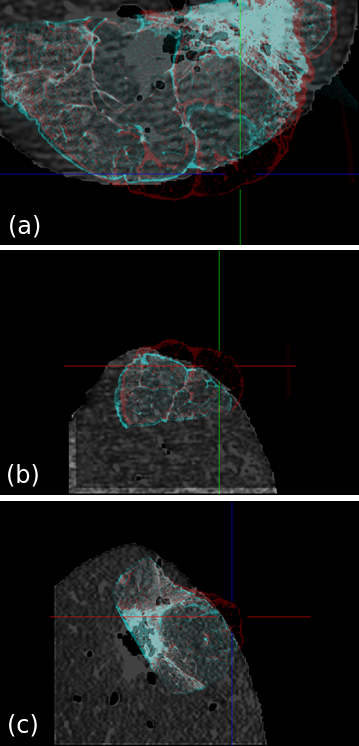}
   \caption{Alignment of \uCT{} lung specimen with pre-operative clinical CT in axial (a), coronal (b), and sagittal (c) planes:  before registration (red), after non-rigid alignment (cyan).}
   \label{fig:axcosa}
\end{figure}
In order to quantitatively evaluate the registration performance, we measure the average minimum surface distance (AvgDist) between the lung surface extracted from clinical CT and \uCT{} before and after non-rigid registration. Figure \ref{fig:surface} shows the extracted surfaces used for measurement before and after non-rigid alignment. The AvgDist measures are given in Table \ref{tab:min_dist}. A reduction from 3.3 $\pm$ 2.9 (range: [0.1, 15.9]) to 2.3 mm $\pm$ 2.8 (range: [0.0, 15.3]) mm on average can be observed. This is a significant improvement with p $<$ 0.001 (Wilcoxon signed rank test). 
\begin{figure}[htb]
  \centering
   \includegraphics[width=\figscale\columnwidth, clip]{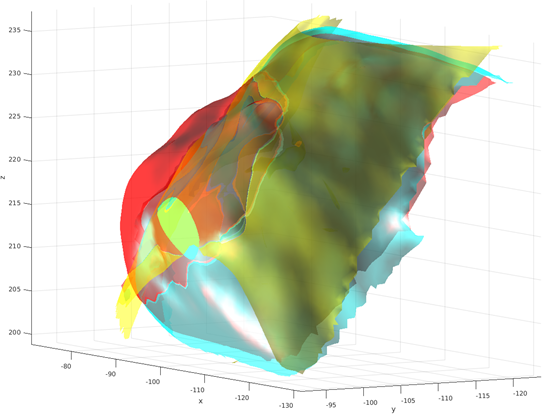}
   \caption{Surface alignment of \uCT{} lung specimen with pre-operative clinical CT (yellow): before registration (red), after non-rigid alignment (cyan).}
   \label{fig:surface}
\end{figure}
\begin{table} [htb]
\begin{center}
\caption{Average minimum distances (AvgDist) in mm between the lung surface extracted from clinical CT and \uCT{} before and after non-rigid registration.}
\label{tab:min_dist}
\footnotesize
  \begin{tabular}{l|rrrr|rrrr}
    \hline
    \hline    
       \textbf{AvgDist} & \multicolumn{4}{c}{\textbf{before registration}} & \multicolumn{4}{c}{\textbf{after registration}}\\
    \textbf{[mm]} & mean & std & min. & max. & mean & std & min. & max. \\ 
    \hline
    case 1 & 3.6 & 3.1 & 0.0 & 17.2             & 2.6  & 2.8 & 0.0   & 15.9  \\
    case 2 & 2.9 & 2.7 & 0.1 & 14.6             & 2.0  & 2.7 & 0.0   & 14.7   \\   
    \textbf{mean} & \textbf{3.3}   & \textbf{2.9} & \textbf{0.1} & \textbf{15.9}
                                 & \textbf{2.3}  & \textbf{2.8} & \textbf{0.0}   & \textbf{15.3}   \\
    \hline
    \hline
  \end{tabular}
\end{center}
\end{table}
\section{CONCLUSIONS}
\label{sec:conclusions}
\noindent We presented a method for non-rigid alignment between images of pre-operative clinical CT and x-ray microtomography (\uCT) of lung pathology. After manual initialization using a few landmark points and affine alignment, several levels of non-rigid registration between down-sampled (in the case of  \uCT{}) and up-sampled (in the case of clinical CT) representations of the image are performed. This allows us to recover the volume differences introduced by the resection and preparation of the lung specimen. The proposed multi-scale image fusion approach will allow further registration with even finer resolutions of \uCT{} (up to 10 $\mu$m resolution) \cite{nagara2016cascade} and ultimately with histopathological microscopy images for further macro to micro image fusion that can aid medical image analysis.
\vspace{1em}
\section*{ACKNOWLEDGEMENTS}
\noindent This paper was supported by MEXT KAKENHI (26108006 and 15H01116).
\vspace{1em}
\bibliographystyle{ieeetr}  
\bibliography{references_ifmia2017}

\end{document}